\documentclass{article}

\usepackage{arxiv}

\usepackage[utf8]{inputenc} 
\usepackage[T1]{fontenc}    
\usepackage{hyperref}       
\usepackage{url}            
\usepackage{booktabs}       
\usepackage{amsfonts}       
\usepackage{nicefrac}       
\usepackage{microtype}      
\usepackage{lipsum}
\usepackage{graphicx}
\usepackage{algorithm}
\usepackage{algpseudocode}
\usepackage{pgfplots}
\usepackage{pgfplotstable}
\pgfplotsset{compat=1.18}
\graphicspath{ {./images/} }

\title{Advancing Eurasia Fire Understanding Through Machine Learning Techniques}

\author{
 Kriuk Boris \\
  Department of Computer and Electronic Engineering\\
  Hong Kong University of Science and Technology\\
  \texttt{bkriuk@connect.ust.hk} \\
}

\begin{document}
\maketitle
\begin{abstract}
Modern fire management systems increasingly rely on satellite data and weather forecasting; however, access to comprehensive datasets remains limited due to proprietary restrictions. Despite the ecological significance of wildfires, large-scale, multi-regional research is constrained by data scarcity. Russia’s diverse ecosystems play a crucial role in shaping Eurasian fire dynamics, yet they remain underexplored. This study addresses existing gaps by introducing an open-access dataset that captures detailed fire incidents alongside corresponding meteorological conditions. We present one of the most extensive datasets available for wildfire analysis in Russia, covering 13 consecutive months of observations. Leveraging machine learning techniques, we conduct exploratory data analysis and develop predictive models to identify key fire behavior patterns across different fire categories and ecosystems. Our results highlight the critical influence of environmental factor patterns on fire occurrence and spread behavior. By improving the understanding of wildfire dynamics in Eurasia, this work contributes to more effective, data-driven approaches for proactive fire management in the face of evolving environmental conditions.
\end{abstract}


\section{Introduction}
Fire is a natural phenomenon that plays a dual role in ecosystems, acting both as a vital ecological process and a destructive force. With its diversity and scale, the ecosystem of Russian territories is crucial for understanding the processes in Eurasia, and the research conducted on these regions is insufficient. In Russia, a country characterized by vast landscapes ranging from dense forests to tundra, wildfires have become increasingly prevalent, particularly in the context of  anthropogenic activities. The complexity of fire dynamics, influenced by factors such as weather conditions, vegetation type, and human interaction, necessitates advanced methods for understanding and managing such events proactively. 

Despite the ecological role of fire [1,2], the challenges associated with wildfire management are profound. Russia has experienced a rise in wildfire incidents, exacerbated by climatic shifts, such as rising temperatures and prolonged droughts. Such incidents not only lead to environmental degradation but also contribute to significant economic losses and numerous public health issues. Traditional fire management practices often rely on historical data and expert knowledge, which may not adequately capture the complex and dynamic nature of fire behavior in the face of changing environmental conditions.

Modern fire management systems increasingly rely on satellite data and weather forecasting to improve predictive capabilities of wildfire monitoring [3]. By utilizing remote sensing technology, these systems can monitor environmental conditions in real-time, providing insights into factors such as vegetation moisture levels, temperature fluctuations, and wind patterns. However, a significant challenge arises from the limited accessibility of this information. Much of the satellite data and weather forecasts that could be instrumental in improving fire prediction models are often kept hidden or released in incomplete formats by companies developing advanced systems. This lack of transparency is primarily driven by competitive concerns, as organizations seek to safeguard their proprietary algorithms and data sources. Consequently, the restricted availability of comprehensive datasets hampers collaboration among researchers and practitioners, ultimately limiting the effectiveness of fire management strategies that could benefit from a more open exchange of information.

To overcome the challenges posed by limited access to comprehensive fire data, we utilized various weather API resources to compile an extensive dataset covering 13 consecutive months of wildfire incidents across Russia. This effort has resulted in one of the most detailed region-specific datasets publicly available, encompassing critical variables such as fire location and the corresponding associated meteorological conditions. By leveraging this data, we performed initial analysis using machine learning techniques to identify patterns that inform fire behavior. The findings enabled us to draw conclusions about the factors influencing wildfires of various types in different continent regions and ecosystems and build predictive models, contributing valuable knowledge to improve existing open source fire management practices.

\section{Literature Review}

\begin{figure}
\begin{center}
\includegraphics[height=9cm]{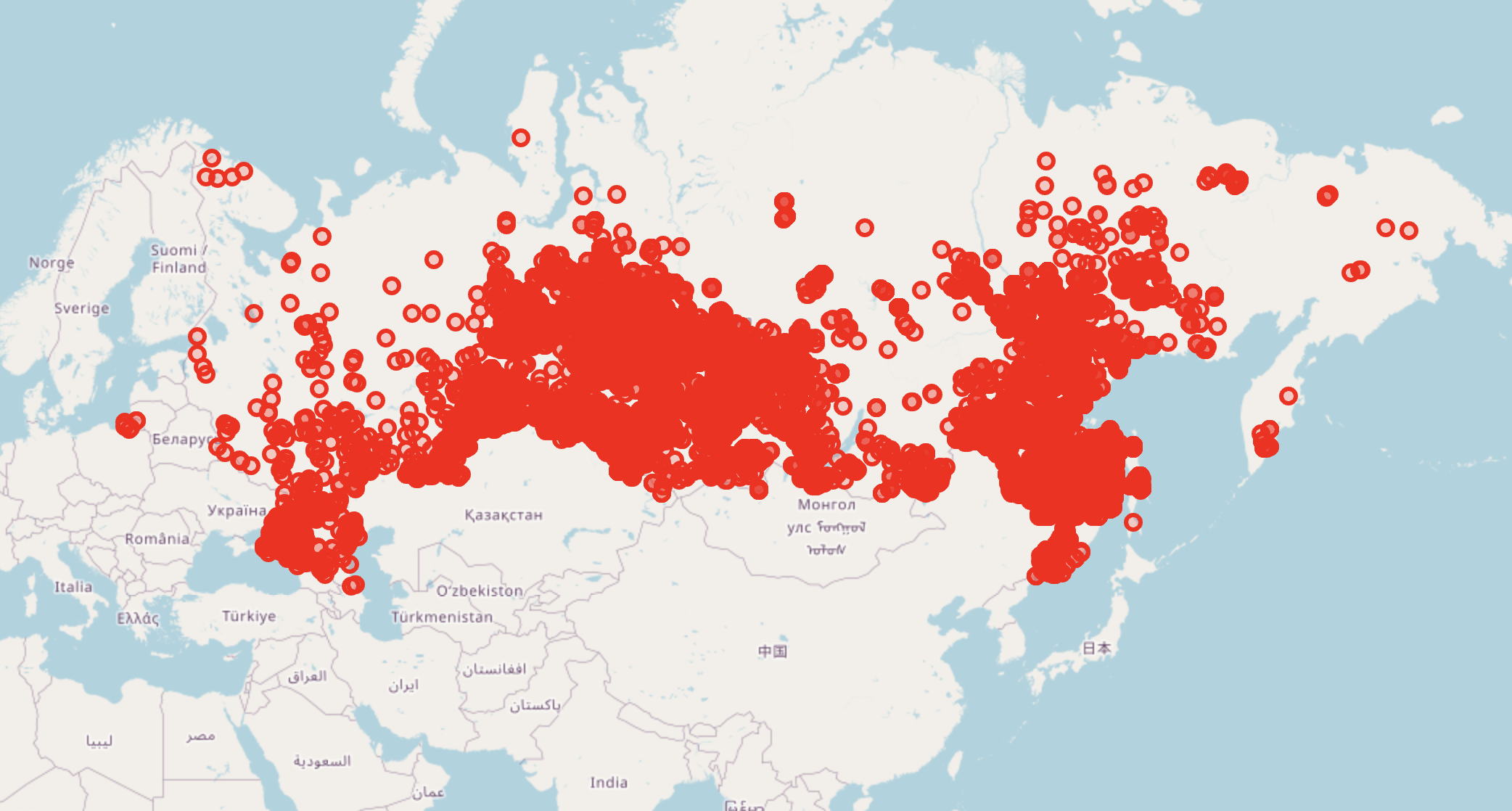}
\end{center}
\caption 
{ \label{fig1}
Unique Fire Locations.} 
\end{figure} 

The foundation of fire prediction modeling was built on statistical approaches, primarily using logistic regression and historical fire occurrence data [4,5,6]. These methods, while fundamental, often struggled to capture the complex, non-linear relationships inherent in fire behavior. Early works by Preisler [7,8] established baseline methodologies using generalized linear models, which, though limited, provided the scaffolding for more sophisticated approaches. The initial statistical frameworks employed analysis for binary prediction outcomes, Poisson regression for count-based fire occurrence modeling, maximum likelihood estimation techniques, and Bayesian inference methods for parameter estimation. These early methods were notably constrained by their limited ability to capture non-linear relationships, difficulty in handling multiple interactive variables, reduced effectiveness with sparse or imbalanced datasets, and the inherent assumption of independence between observations [9,10].

\begin{table}[ht!]
\centering
\caption{Unique Values per Column}
\small
\begin{tabular}{l r}
\toprule
Column Name & Number of Unique Values \\
\midrule
type name              & 5 \\
type id                & 5 \\
lon                    & 25708 \\
lat                    & 23547 \\
temperature (c)         & 2581 \\
precipitation (mm)      & 571 \\
relative humidity     & 4028 \\
wind speed (ms)         & 532 \\
solar radiation       & 731 \\
\bottomrule
\end{tabular}
\label{tab:unique_values}
\end{table}

The transition to machine learning marked a significant advancement in prediction capabilities. Ensemble methods emerged as powerful tools, with Random Forests leading the transformation [11]. Studies [12,13] demonstrated how Random Forests could effectively process high-dimensional feature spaces while maintaining interpretability, a crucial factor for practical implementation. The later works showcased superior performance in feature importance ranking, handling missing data, managing high-dimensional feature spaces, and maintaining strong performance across diverse geographical regions. The development of gradient boosting methods, particularly XGBoost and LightGBM implementations, further enhanced the field by providing improved computational efficiency on structured weather data, better handling of imbalanced datasets, enhanced feature selection capabilities, and superior cross-validation performance [14,15,16].

The paradigm shift toward deep learning has revolutionized fire prediction capabilities. Convolutional Neural Networks (CNNs) have proven particularly effective in processing satellite imagery and spatial data. Notable work by Zhang et al. [17] demonstrated how multi-layer CNNs can integrate multiple data streams, from meteorological conditions to vegetation indices, achieving prediction accuracies exceeding 85 percent. Their implementation demonstrated high-level performance across diverse landscapes, effective feature extraction from complex spatial data, and improved generalization across different geographical regions. Long Short-Term Memory (LSTM) networks have excelled in capturing temporal dependencies in fire progression, with recent implementations showing particular promise in predicting fire spread patterns over extended periods [18,19]. Such networks have advanced capabilities in sequential pattern recognition, long-term dependency modeling, dynamic time series analysis, and adaptive learning rates for varying time scales.

Contemporary research has focused on incorporating both spatial and temporal dimensions simultaneously through hybrid approaches. Graph Neural Networks (GNNs) have emerged as powerful tools for modeling spatial relationships between different regions, while attention mechanisms help focus on critical temporal patterns [20,21]. GNNs have demonstrated exceptional capabilities in spatial relationship modeling, network-based fire spread prediction, regional interdependence analysis, and topological feature extraction. The attention mechanisms have further enhanced predictions through dynamic feature weighting, temporal pattern recognition, multi-scale temporal analysis, and context-aware prediction frameworks.

The operational requirements of fire prediction have driven research into efficient computing approaches. Edge computing solutions have been developed to process satellite data and sensor inputs in real-time, enabling rapid response capabilities. These developments include real-time data processing capabilities, distributed computing architectures, low-latency prediction systems, and resource-optimized deployment strategies [22]. System integration advances have focused on sensor network integration, satellite data processing pipelines, multi-source data fusion frameworks, and scalable computing infrastructure [23].

Emerging areas of investigation in fire prediction modeling include explainable AI frameworks, transfer learning applications, multi-modal integration, and computational efficiency improvements [24]. Research in explainable AI focuses on model interpretation methods, uncertainty quantification, confidence scoring systems, and decision support integration. Transfer learning applications explore cross-region model adaptation, domain-specific fine-tuning, feature transfer optimization, and model portability enhancement. Multi-modal integration advances investigate sensor fusion techniques, data synchronization methods, cross-platform compatibility, and real-time data integration. Computational efficiency research examines model compression techniques, inference optimization, resource allocation strategies, and distributed computing frameworks [25,26]. 

\section{Methodology}
\subsection{Dataset Overview}
In this research paper section, we present a comprehensive dataset FiresRu focused on fire events and their associated meteorological conditions in Russia tracked for 13 consecutive months. The dataset comprises multiple parameters recorded for various types of fire events, including natural fires, forest fires, controlled burns, uncontrolled burns, and peat fires.

The dataset structure includes temporal information (dt), categorical classification of fire types (type name and type id), geographical coordinates (longitude and latitude), and five important meteorological parameters: temperature in Celsius, precipitation in millimeters, relative humidity as a percentage, wind speed in meters per second, and solar radiation.

Our dataset, with details demonstrated in Table 1, contains 26,681 unique observations spanning across a significant geographical area, with longitudes ranging from 20.37°E to 175.87°E and latitudes from 41.71°N to 70.33°N. This extensive coverage suggests the data encompasses a large portion of the territory, as shown in Figure 1.

The temperature measurements in the dataset show considerable variation, ranging from -34.88°C to 31.94°C, with a mean temperature of approximately 17.89°C. Such a wide range reflects the diverse climatic conditions across the studied region and different seasons. The median temperature of 19.08°C suggests that many fire events were recorded during warmer periods, which is consistent with typical fire season patterns.

Precipitation data reveals relatively dry conditions during most recorded events, with a mean of 0.46mm and a median of 0.04mm. The maximum recorded precipitation was 25.85mm, though the 75th percentile at 0.30mm indicates that most fires occurred during periods of minimal precipitation, which is logical given the nature of fire events.

The fire events are categorized into five distinct types, represented by both type name and numerical identifiers (type id). The data points reveal an uneven distribution among fire types, introducing a typical class imbalance challenge.

The FiresRu dataset provides valuable research information and becomes one of the most detailed existing resources for studying the relationship between meteorological conditions and different types of fire events in the region. The comprehensive nature of the meteorological parameters included allows for analysis of the environmental conditions associated with various fire types, which is crucial for fire prediction and management strategies.

\subsection{Data Analysis}

\begin{figure}
\begin{center}
\includegraphics[height=11cm]{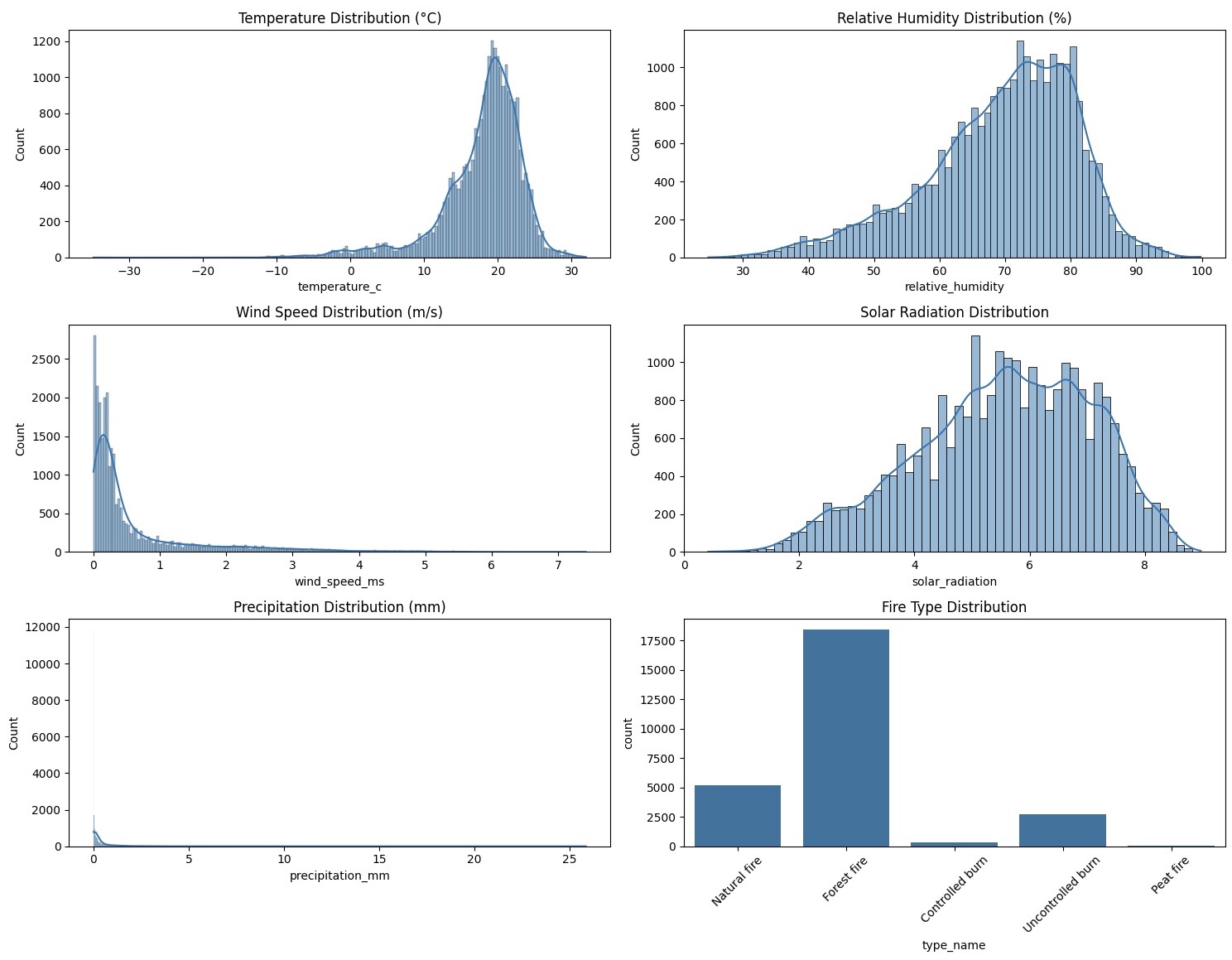}
\end{center}
\caption 
{ \label{fig1}
Feature Values Distribution.} 
\end{figure} 

\begin{figure}
\begin{center}
\includegraphics[height=11cm]{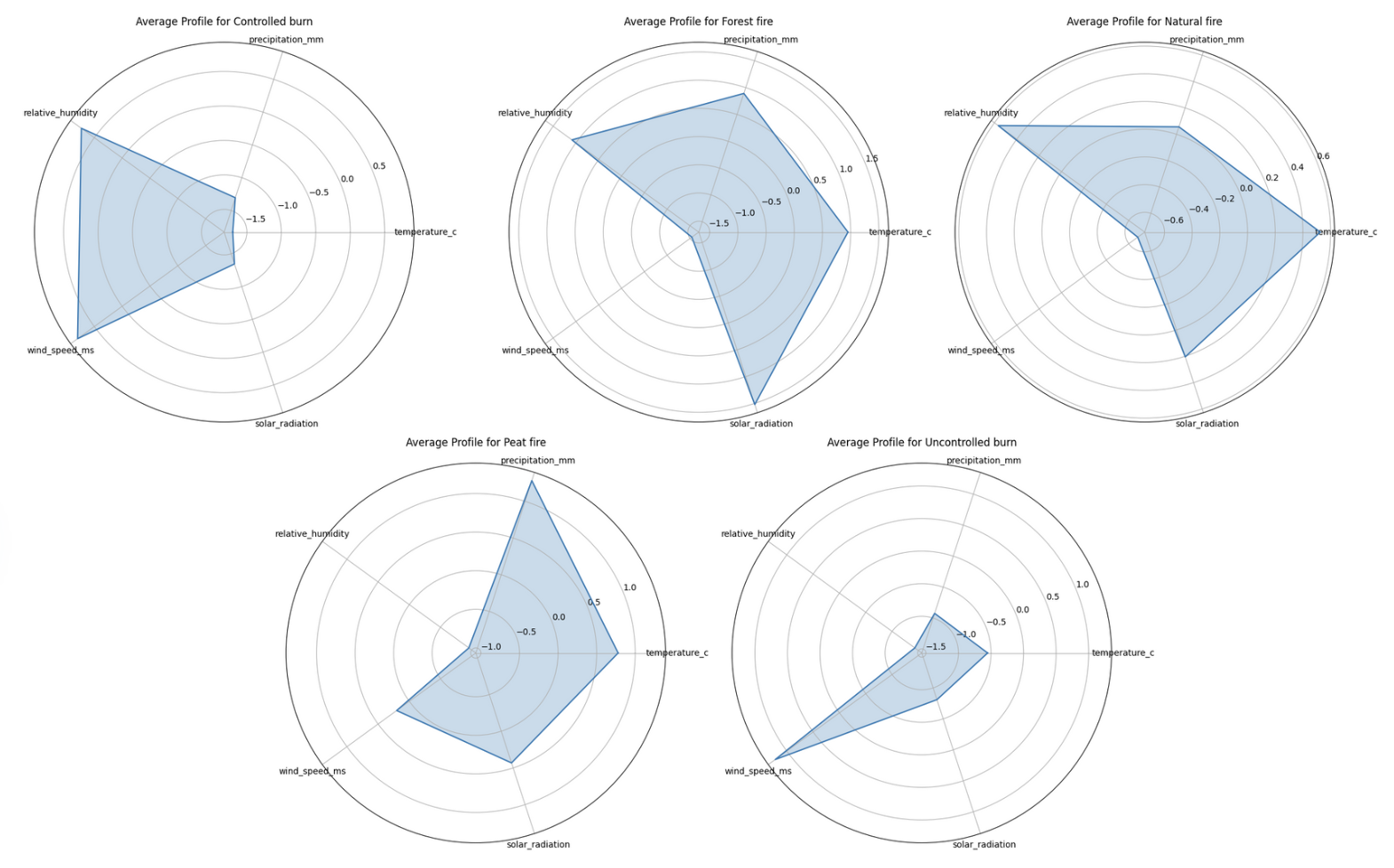}
\end{center}
\caption 
{ \label{fig1}
Average Profiles of Fire Types.} 
\end{figure} 

As shown in Fig. 2, the temperature pattern with a peak around 20°C indicates that most fires occur during moderate temperature conditions, not necessarily during extreme heat. This observation highlights that fires are not primarily driven by high temperatures alone. The distribution suggests that other factors like fuel availability and moisture content may be even more critical than absolute temperature for certain wildfire cases.

The humidity data is particularly interesting - the peak at 70-80 percent relative humidity might seem counterintuitive, as fires are often associated with dry conditions. We associate such humidity levels with the idea that many fires start during morning or evening hours when humidity is higher, or reflect conditions in forested areas where ambient humidity remains elevated even during fire events, like swamps and peatbogs.

Wind speed distribution shows that most fires occur during relatively calm conditions (0-2 m/s). While high winds can exacerbate fire spread, the data suggests that initial fire occurrence doesn't require strong winds. 

Solar radiation shows a bimodal pattern, corresponding to various geographical locations throughout the year, and the precipitation pattern is demonstrating that most fires occur during periods with minimal rainfall. The sharp drop-off above 1mm suggests this might be a natural threshold for fire suppression at the beginning stages, making it a logical barrier for wildfire uncontrolled spread.

The fire type categorization heavily favors forest fires, followed by natural fires, highlighting previous statements regarding class imbalance present in the dataset.

\begin{figure}
\begin{center}
\includegraphics[height=7cm]{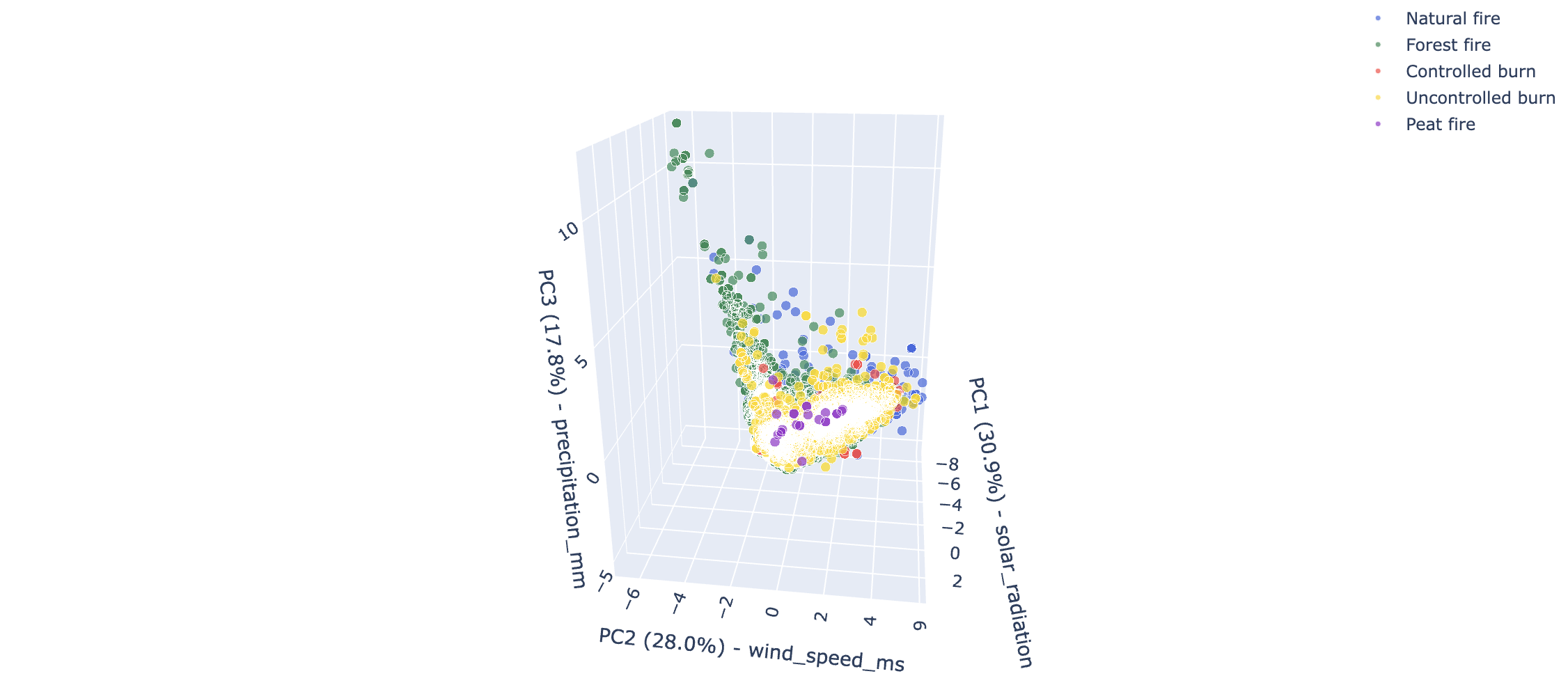}
\end{center}
\caption 
{ \label{fig1}
Fire Type Clusters in 3D PCA Space.} 
\end{figure} 

We scale features with Z-score normalization and analyze the profile of each fire type separately, as shown in Fig. 3. Our observations indicate that wind speed emerges as the most significant parameter for controlled and uncontrolled burns, demonstrating a strong deviation from baseline conditions. Based on the findings, lower humidity levels and higher temperatures contribute to the burn becoming uncontrolled. This relationship may be attributed to the fact that stronger wind speeds create more aggressive conditions for fire spread, although such a pattern is specifically related to fire occurrence rather than to the established subsequent fire behavior.

Unlike for burns, temperature appears to play a notable role and have a strong influence on natural and forest fires occurrence. We conclude that burns tend to initiate during relatively cooler conditions, a finding that may seem counterintuitive given the traditional association between high temperatures and fire risk. However, this behavior may indicate clear weather patterns that increase the likelihood of outdoor activities among the general public, which is a common cause of a burn occurrence risk.

The predominant factors influencing natural and forest fires success include high solar radiation and relative humidity. This behavior can be explained by the way solar radiation directly affects the moisture content of potential fuel sources. When solar radiation levels are high, it leads to increased surface temperatures and accelerated evaporation rates, creating conditions more conducive to fire ignition and spread. The relationship between relative humidity and fire behavior is complex, as it affects both the moisture content of fuels and the overall fire environment. Lower relative humidity typically results in drier fuels, making them more susceptible to ignition and rapid combustion.

Still, forest fires are stronger shaped by higher temperatures, since the wood is extremely susceptible to burning. The thermal properties of wood, particularly its cellulose and lignin components, make it highly reactive to heat. As temperatures rise, these organic compounds begin to break down through pyrolysis, releasing volatile gases that further fuel the combustion process. The density and arrangement of forest fuels, combined with elevated temperatures, create a self-reinforcing cycle that can lead to intense and rapidly spreading fires.

At the same time, natural fires are promoted in environments with higher humidity levels. Such pattern shows an interesting divergence from forest fire behavior and can be attributed to several factors. In environments with higher humidity, there is often more abundant vegetation growth, which provides increased fuel loads. Additionally, humid conditions can lead to more frequent lightning strikes, a primary natural ignition source. The presence of moisture in the air can also influence fire behavior through convection processes and the formation of pyrocumulus clouds, which can create their own weather patterns and affect fire spread dynamics.

Analysis of fire type patterns reveals distinct meteorological signatures across fire categories, as shown in Figure 4. Principal Component Analysis (PCA) identified three primary components explaining 76.7 percent of total variance in the meteorological conditions associated with different fire types.

The first principal component (PC1), explaining 30.9 percent of variance, is primarily characterized by a strong positive correlation with solar radiation (0.603) and temperature (0.485), and negative correlations with relative humidity (-0.457) and precipitation (-0.388). Such pattern aligns with findings from Jolly et al. [27] regarding global fire weather trends, where increased fire activity correlates with warmer, drier conditions. PC2 accounts for 28.0 percent of variance and is notably dominated by wind speed (0.699), with strong negative loading from relative humidity (-0.505). This component likely represents atmospheric conditions conducive to fire spread, consistent with studies by Rothermel [28] on fire behavior modeling. PC3 (17.8 percent variance) shows strong positive correlations with precipitation (0.695) and temperature (0.489), while maintaining negative correlation with relative humidity (-0.414). We believe this component represents seasonal transitional periods or specific meteorological conditions associated with certain fire types, particularly relevant to controlled burns as documented in prescribed fire management literature [29].

\subsection{Modelling}

We scale the dataset features, normalize and use for ML algorithms’ training to predict fire type based on input data. We split the data as 80-10-10 for training, validation, and testing. The analysis of wildfire prediction models reveals interesting patterns in classification performance across different algorithms, as shown in Table 2. The Extra Trees classifier emerged as the top performer with an accuracy of 88.34 percent and a weighted F1 score of 0.8791, closely followed by Random Forest at 88.23 percent accuracy, both demonstrating a high-level capability in handling the complex relationships between environmental variables and fire occurrences. The superior performance can be attributed to algorithms' ensemble nature and the ability to handle non-linear interactions between diverse features yet staying generalized.

\begin{table}[ht!]
\centering
\caption{Model Performance Comparison}
\small
\begin{tabular}{l r r r r}
\toprule
Model & Accuracy & Macro Avg F1 & Weighted Avg F1 \\
\midrule
Extra Trees & 0.8834 & 0.7190 & 0.8791 \\
Random Forest & 0.8823 & 0.7217 & 0.8778 \\
Bagging Classifier & 0.8756 & 0.7103 & 0.8727 \\
XGBoost & 0.8688 & 0.6925 & 0.8607 \\
CatBoost & 0.8508 & 0.6530 & 0.8378 \\
LightGBM & 0.8482 & 0.6460 & 0.8334 \\
KNN & 0.8351 & 0.5461 & 0.8290 \\
Gradient Boosting & 0.7909 & 0.5454 & 0.7522 \\
Neural Network MLP& 0.7766 & 0.3943 & 0.7411 \\
SVM & 0.7627 & 0.3502 & 0.7019 \\
Logistic Regression & 0.7376 & 0.3017 & 0.6619 \\
Linear SVM & 0.7290 & 0.2689 & 0.6389 \\
AdaBoost & 0.7264 & 0.3193 & 0.6673 \\
QDA & 0.7170 & 0.3199 & 0.6677 \\
Naive Bayes & 0.7159 & 0.3109 & 0.6577 \\
\bottomrule
\end{tabular}
\label{tab:model_performance}
\end{table}

The second tier of performance is dominated by other ensemble methods, with the Bagging Classifier and XGBoost achieving accuracies of 87.56 percent and 86.88 percent respectively. Such results parallel the ecological concept of resilience through diversity, where multiple perspectives (in this case, different models within the ensemble) contribute to a more adaptive system. The strong performance of these ensemble methods suggests that fire behavior patterns, like many natural phenomena, are best understood through multiple complementary perspectives rather than a single analytical approach.

Gradient boosting variants (CatBoost and LightGBM) demonstrated solid performance with accuracies around 85 percent, while maintaining strong weighted F1 scores above 0.83. Such performance level indicates their ability to capture the nuanced relationships between meteorological conditions and fire occurrence, similar to how ecological succession processes involve incremental improvements and adaptations. The gradient boosting approach also excels in capturing the environmental thresholds that can trigger significant changes in fire behavior.

The K-Nearest Neighbors algorithm achieved a respectable 83.51 percent accuracy, representing a transition point in model performance. This method's relative success demonstrates that fire events often share similar environmental conditions, much like how ecological niches tend to have characteristic environmental parameters. The spatial and conditional proximity principles underlying KNN align well with the geographical and meteorological patterns that influence fire behavior.

A notable performance gap appears with the traditional machine learning approaches, where the Gradient Boosting and Neural Network models achieved accuracies of 79.09 percent and 77.66 percent respectively. The performance drop mirrors the challenges in ecological modeling where simple linear or neural relationships often fail to capture the full complexity of environmental interactions. These models' limitations reflect the non-linear and partially noisy nature of fire behavior in natural systems.

The lower-performing group includes traditional statistical approaches such as SVM (76.27 percent), Logistic Regression (73.76 percent), and Linear SVM (72.90 percent). Their weaker performance underscores the inherent complexity of fire behavior prediction, which, like many ecological processes, rarely follows linear or easily separable patterns. These models' struggles parallel the limitations of reductionist approaches in understanding complex ecological phenomena.

The final tier includes AdaBoost, QDA, and Naive Bayes, with accuracies around 71-72 percent. Despite their sophisticated mathematical foundations, such models' lower performance highlights how fire ecology requires methods that can capture both the broad patterns and fine-grained interactions in highly imbalanced environmental data. The performance limitations of AdaBoost, QDA, and Naive Bayes models highlight the challenges in modeling complex ecological systems with assumptions of statistical independence or quadratic relationships may not hold true.

The feature importance analysis of wildfire risk factors using a Random Forest classifier revealed patterns of the environmental variables that contribute most strongly to fire occurrence and behavior in the region, as shown in Fig. 5.

Solar radiation emerged as the most influential factor, accounting for over 23 percent of the model's predictive power in the built-in metrics and showing the highest permutation importance as well. The observation aligns with established understanding of solar radiation influence on fuel drying and ground temperature conditions, hence, creating favorable conditions for fire ignition and spread.

Temperature proved to be the second most important feature according to the Random Forest's built-in metrics with over 22.5 percent score, though it ranked third in permutation importance. Such discrepancy between the two measurement methods used suggests that while temperature is crucial, its predictive power may be partially correlated with other variables.

Wind speed demonstrated substantial importance, ranking third in built-in metrics (over 21 percent) but second in permutation importance. The finding reinforces a critical role of wind in fire behavior, particularly in fire spread and intensity. The low standard deviation in the permutation analysis indicates the consistency of wind speed's influence across different scenarios.

Relative humidity showed consistent importance across both metrics, highlighting its role in determining fuel moisture content and fire susceptibility. The low standard deviation in permutation testing suggests its influence is stable and fundamental across different conditions.

Precipitation demonstrated the lowest relative importance among the studied variables, though this should not diminish its significance in fire prevention and control. The lower ranking might reflect the more immediate effects of other variables or the temporal nature of precipitation's impact on fire risk.

\begin{figure}
\begin{center}
\includegraphics[height=7cm]{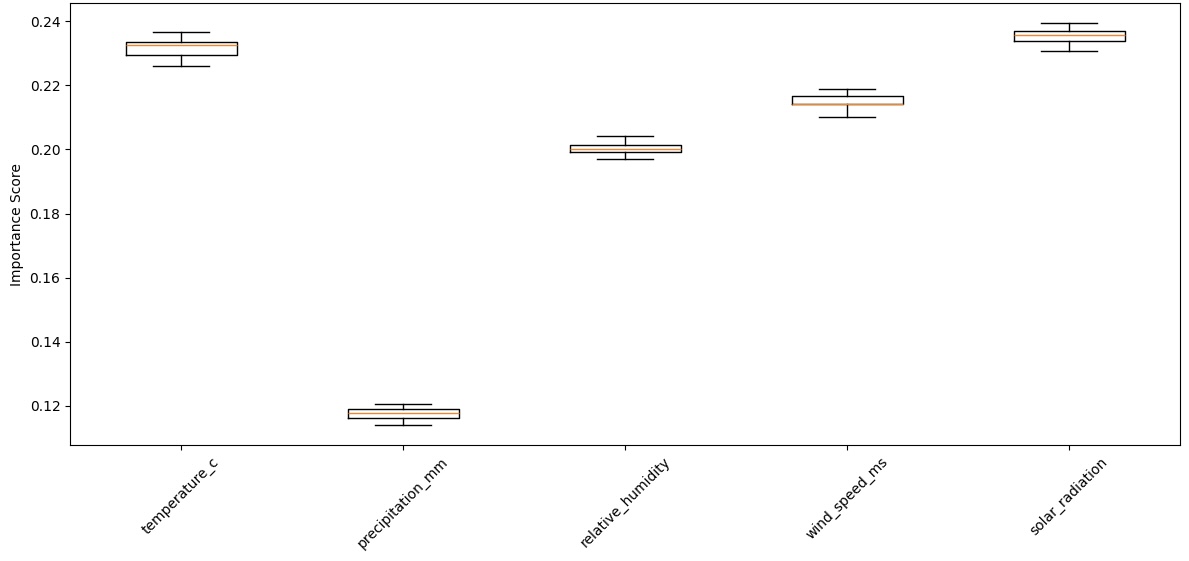}
\end{center}
\caption 
{ \label{fig1}
Feature Importance in Random Forest Model.} 
\end{figure} 

\begin{figure}
\begin{center}
\includegraphics[height=7cm]{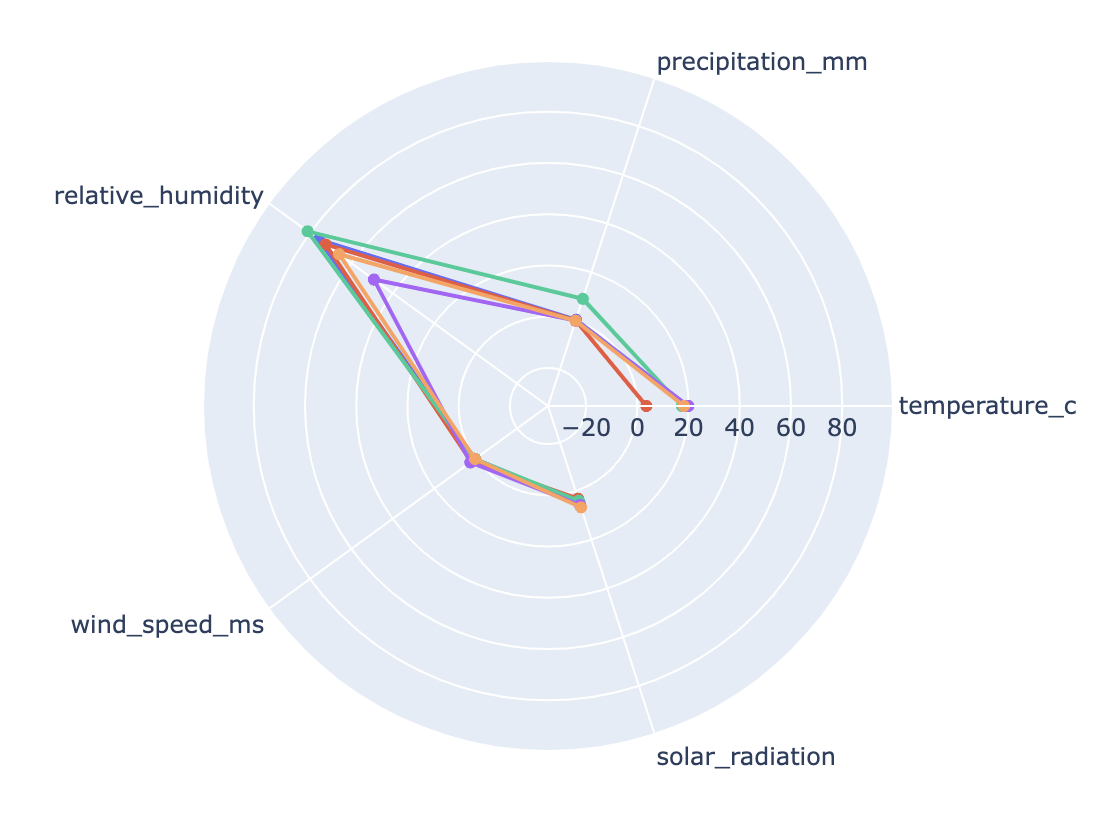}
\end{center}
\caption 
{ \label{fig1}
Cluster Feature Importance Across All Fire Categories.} 
\end{figure} 

The dataset's structure reveals the complexity of fire prediction, encompassing both fine-grained meteorological measurements and broader geographical patterns. The data variability, combined with the spatial components represented by latitude and longitude coordinates, creates a challenging prediction landscape that tree-based ensemble methods appear better equipped to handle compared to other common machine learning approaches.

The analysis of wildfire predictors for Random Forest model (Fig. 5, Fig. 6) reveals a hierarchy of environmental factors, with solar radiation emerging as the most influential variable, accounting for approximately 24 percent of the model's decision-making process. Solar radiation's dominant role stems from its direct impact on fire dynamics through multiple mechanisms: it provides the primary energy input that heats fuel surfaces, drives moisture evaporation from vegetation, and creates the temperature gradients that influence local fire weather. When solar radiation intensifies, it triggers a cascade of fire-promoting conditions - the heat dries out fine fuels like leaves and twigs, bringing them closer to their ignition point, while simultaneously warming the surrounding air. The warming creates unstable atmospheric conditions that can enhance fire behavior through increased vertical air movement and reduced relative humidity. Such relationship is particularly evident in how fire activity often peaks during periods of maximum solar radiation, when fuels have been subjected to hours of direct heating and the atmosphere has become most conducive to rapid fire spread.

\begin{figure}
\begin{center}
\includegraphics[height=10cm]{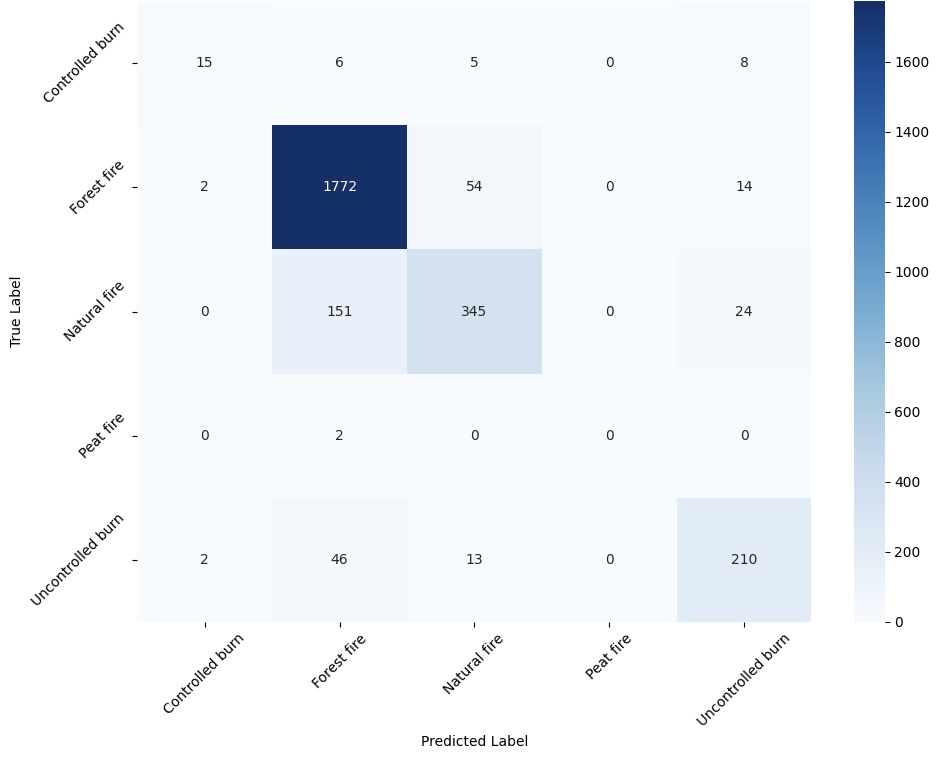}
\end{center}
\caption 
{ \label{fig1}
Confusion Matrix in Random Forest Model.} 
\end{figure} 

Temperature follows closely as the second most important factor, contributing about 23 percent of the predictive power. This significant influence stems from temperature's fundamental role in fire behavior: higher ambient temperatures reduce the energy required to reach ignition point, pre-condition fuels by reducing their moisture content, and create atmospheric conditions conducive to rapid fire spread. The relationship between temperature and fire intensity manifests in multiple ways - warmer conditions accelerate the rate of pyrolysis (the thermal decomposition of vegetation), while also creating convection currents that enhance vertical fire spread and crown fire development. These effects are particularly evident in how afternoon temperature peaks often coincide with the most aggressive fire behavior, when fuel temperatures reach their maximum and the atmosphere becomes most unstable, leading to the characteristic "blow-up" conditions that firefighters particularly fear.

Wind speed emerges as the third most significant factor, accounting for roughly 21 percent of the model's predictive capability. Such prominence reflects wind's critical role in fire behavior through three primary mechanisms: it supplies fresh oxygen to the combustion zone, pushes flames toward new fuel sources, and promotes pre-heating of potential fuel through the transport of hot air and embers. Wind acts as a natural bellows system, intensifying fires by creating a feedback loop where increased oxygen supply leads to more intense burning, which in turn generates stronger fire-induced winds. The dynamic relationship between wind and fire behavior is particularly evident in phenomena like fire whirls and crown fires, where wind patterns can transform a ground-level fire into an intense aerial conflagration, jumping between tree canopies at speeds significantly higher than ground-based spread rates.

Relative humidity and precipitation together account for the remaining 32 percent of the model's decision-making process, with humidity at about 20 percent and precipitation at 12 percent. This moisture-related hierarchy reflects how atmospheric and ground moisture interact with fire behavior through distinct mechanisms. Relative humidity directly influences how quickly fuels can dry out or absorb moisture from the air - when humidity drops, fine fuels like dead grass and leaf litter can rapidly lose moisture and become highly flammable within hours. Precipitation's lower predictive importance suggests that while rainfall events temporarily suppress fire risk, their effects are less persistent than other factors. This aligns with field observations where severe fires can occur shortly after rain if other conditions are extreme, particularly when high temperatures and strong winds rapidly dry out surface fuels. The interaction between these moisture variables helps explain why some regions can experience high fire danger even with recent rainfall history - it's the combination of low humidity, high temperatures, and strong winds that creates the most hazardous fire conditions, regardless of recent precipitation patterns.

The confusion matrix analysis (Fig. 7) demonstrates the Random Forest model's classification performance across distinct fire categories. The confusion matrix highlights several significant patterns of misclassification that can be explained through the lens of environmental conditions and fire behavior characteristics. The most notable misclassifications occurred between forest fires and controlled burns (151 cases), which is understandable given that both types share similar meteorological prerequisites. Such confusion likely occurs due to controlled burns intentional initiation under weather conditions that typically support forest fires - moderate temperatures, manageable wind speeds, and appropriate humidity levels - to ensure controlled spread and effective management.

The misclassification of 46 peat fires as forest fires and 13 as controlled burns can be attributed to the complex nature of peat fire behavior. While peat fires are distinct in their subsurface burning characteristics, the surface-level meteorological conditions (temperature, wind speed, solar radiation) may mirror those of forest fires, particularly during the initial stages. The model's reliance on above-ground weather parameters may limit its ability to distinguish the unique characteristics of peat fires [30,31], which often persist under conditions that would not typically support surface fires.

Natural fires were frequently misclassified (6 as forest fires, 5 as controlled burns, and 8 as peat fires), suggesting that the meteorological signatures of naturally ignited fires overlap significantly with other fire types. This is logical given that natural fires can evolve into any of these other categories depending on the environmental conditions and available fuel types. The relatively high solar radiation importance (23.86 percent) in the model may explain some of these misclassifications, as high solar radiation and temperature conditions can create similar fire-conducive environments across multiple fire types.

The complete failure to correctly identify uncontrolled burns is particularly noteworthy. We conclude that severe class imbalance and the nature of uncontrolled burns behavior [32,33], occurring across a wide spectrum of weather conditions, makes them difficult to distinguish based on meteorological parameters. The rapid evolution and unpredictable nature of uncontrolled burns means they may not have a distinct meteorological signature, as they can occur under various weather conditions that would typically be associated with other fire types.

\section{Conclusion}

In this research work we present one of the most detailed region-specific publicly available datasets covering 13 consecutive months of distinct fire occurrences under specific meteorological conditions to increase the understanding level of diverse Eurasian wildfire landscape. Through the application of advanced machine learning techniques, we uncover patterns in fire behavior and prediction. Our analysis demonstrates that ensemble methods, particularly Extra Trees and Random Forest classifiers, achieve superior performance in fire type prediction. The success highlights complex, non-linear nature of fire behavior and the importance of considering multiple environmental factors simultaneously. The feature importance analysis reveals that solar radiation, temperature, and wind speed are the most influential factors in determining fire occurrence and type. We describe hierarchy of environmental factors to provide valuable information for fire management strategies and risk assessment protocols developed by the research groups worldwide. Our study uncovers distinct meteorological signatures for different fire types, though with some notable overlap, particularly between forest fires and controlled burns. The challenges in distinguishing uncontrolled burns from other fire types emphasize the complex nature of fire behavior and the limitations of current prediction models. 

Building upon our analysis of wildfire patterns, several promising research directions emerge. First, the notable overlap between forest fires and controlled burns suggests the need to incorporate additional discriminative features beyond meteorological parameters. Future studies should consider integrating satellite-derived fuel moisture indices, vegetation density metrics, and terrain characteristics to enhance classification accuracy. The complete inability to identify uncontrolled burns highlights a critical area for improvement, potentially addressable through the inclusion of temporal progression data and fire spread parameters.

Our dataset's unique temporal coverage of 13 consecutive months provides an excellent foundation for developing seasonal prediction models. Future research should extend this temporal window to capture multi-year climate patterns and their influence on fire behavior. The strong performance of ensemble methods encourages exploration of hybrid models that could combine the strengths of different machine learning approaches with physical fire behavior models. Such focus could particularly enhance the prediction of peat fires, which showed moderate classification accuracy but significant confusion with other fire types.

The identified hierarchy of environmental factors opens avenues for developing region-specific fire danger rating systems. Future work should focus on validating these relationships across different Eurasian ecological zones and seasons. Additionally, the integration of high-resolution weather forecast data can transform our classification models into a complete predictive early warning system.

\section{Data Availability}
The dataset is available at: https://github.com/sparcus-technologies/FiresRu

\section{Disclosures}

All the authors declare no conflict of interests.

\bibliographystyle{unsrt}  


\end{document}